\documentclass[conference]{IEEEtran}

\usepackage{enumerate}
\usepackage{graphicx}
\usepackage{epsf}
\usepackage{subfigure}
\usepackage{amsmath}
\usepackage{amssymb}
\usepackage{array}
\usepackage{setspace}
\usepackage[amsmath,thmmarks]{ntheorem}
\usepackage[ruled,lined]{algorithm2e}
\usepackage{algorithmic}
\usepackage{color}
\usepackage{amsfonts}
\usepackage{dsfont}
\usepackage{xfrac}
\usepackage{breqn}
\usepackage{bbm}
\usepackage{cite}

\graphicspath{{Fig/},{fig/}}

\theoremseparator{.}

\begin{document}

\title{\huge Federated Deep Reinforcement Learning for Energy Efficient Multi-Functional RIS-Assisted Low-Earth Orbit Networks}

\author{\IEEEauthorblockN{Li-Hsiang Shen, Jyun-Jhe Huang, Kai-Ten Feng$^\dagger$, Lie-Liang Yang*, and Jen-Ming Wu$^\ddagger$}\\
\IEEEauthorblockA{\small
	Department of Communication Engineering, National Central University, Taoyuan, Taiwan\\
	$^\dagger$Department of Electronics and Electrical Engineering, National Yang Ming Chiao Tung University, Hsinchu, Taiwan\\ 
	*Next Generation Wireless, University of Southampton, Southampton, UK\\
	$^\ddagger$Next-generation Communications Research Center, Hon Hai Research Institute, Taipei, Taiwan\\
Email: shen@ncu.edu.tw, kenneth900912@g.ncu.edu.tw, ktfeng@nycu.edu.tw, lly@ecs.soton.ac.uk, and jen-ming.wu@foxconn.com}}

\maketitle

\begin{abstract}
In this paper, a novel network architecture that deploys the multi-functional reconfigurable intelligent surface (MF-RIS) in low-Earth orbit (LEO) is proposed. Unlike traditional RIS with only signal reflection capability, the MF-RIS can reflect, refract, and amplify signals, as well as harvest energy from wireless signals. Given the high energy demands in shadow regions where solar energy is unavailable, MF-RIS is deployed in LEO to enhance signal coverage and improve energy efficiency (EE). To address this, we formulate a long-term EE optimization problem by determining the optimal parameters for MF-RIS configurations, including amplification and phase-shifts, energy harvesting ratios, and LEO transmit beamforming. To address the complex non-convex and non-linear problem, a federated learning enhanced multi-agent deep deterministic policy gradient (FEMAD) scheme is designed. Multi-agent DDPG of each agent can provide the optimal action policy from its interaction to environments, whereas federated learning enables the hidden information exchange among multi-agents. In numerical results, we can observe significant EE improvements compared to the other benchmarks, including centralized deep reinforcement learning as well as distributed multi-agent deep deterministic policy gradient (DDPG). Additionally, the proposed LEO-MF-RIS architecture has demonstrated its effectiveness, achieving the highest EE performance compared to the scenarios of fixed/no energy harvesting in MF-RIS, traditional reflection-only RIS, and deployment without RISs/MF-RISs.

\end{abstract}

\section{Introduction}

In revolutionary era of information explosion, global communication technology is rapidly advancing from fifth-generation (5G) to six-generation (6G), driving increasing demands for high-coverage and high-performance networks \cite{1}. To meet these growing requirements, reconfigurable intelligent surface (RIS) technology has emerged as a promising solution \cite{2}. By adjusting the configuration of RIS elements, a virtual line-of-sight (LoS) link can be established to bypass obstacles between the transmitter and receiver \cite{3}. Due to these advantages, RIS technology can enable precise signal distribution, significantly improving communication performance \cite{my1}. 

Despite its potential, RIS technology still faces challenges, such as half-space coverage limitations and reliance on external power sources, which restrict its full capabilities. To address these issues, the concept of multi-functional RIS (MF-RIS) has been introduced \cite{6}. Unlike traditional RIS, which depends on external power to reflect and manipulate signals, MF-RIS can harvest energy from radio-frequency (RF) signals, enabling self-sufficient operation with fewer needs of batteries or grid power. This energy harvesting capability improves the energy efficiency (EE) and sustainability of MF-RIS. Additionally, MF-RIS extends beyond basic signal reflection by supporting signal amplification, allowing it to enhance both coverage and signal strength when necessary.

With advancements in aerospace technology, satellite communication has demonstrated great potential for achieving global coverage \cite{7}. However, the rapid increase in the number of connected ground users presents significant challenges, particularly in terms of limited bandwidth and energy allocation \cite{8}. Low-Earth orbit (LEO) satellite communication systems have emerged as a key focus due to their advantages of global connectivity and high throughput \cite{9}. LEO systems offer comparatively lower latency than other satellite systems, enabling more efficient real-time communications. However, the long distance involved in LEO satellite communication still results in significant pathloss. Recently, numerous studies have focused on integrating RIS into satellite systems to enhance signal transmission range \cite{10,11}. Additionally, LEO satellites rely on solar panels to harvest energy in sunlit regions. By contrast, in shadow regions they must depend on battery storage for operation. However, most of existing studies have not considered the practical power consumption as well as battery storage of LEO satellites and RIS. To address the issue of insufficient energy, this work proposes the deployment of MF-RIS on LEO satellites, coordinating with solar panels to optimize system energy efficiency.

In this work, we focus on the maximization of EE of the conceived LEO-MF-RIS architecture. However, the rapid variation in channel conditions due to the movement of LEO satellites along their orbits creates a highly complex, challenging problem. To address this, deep reinforcement learning (DRL) is employed to adaptively adjust the network policies in response to the dynamic environment. Moreover, we utilize multi-agent deep deterministic policy gradient (MADDPG) to handle the high-dimensional state space and complex action scenarios. MADDPG adopts an actor-critic (AC) framework, enabling each agent to determine the optimal action policy through interaction with the environment \cite{my2}. To further enhance cooperative policy decision-making, federated learning (FL) is incorporated to facilitate exchange and aggregation of model parameters among LEO agents for improved information sharing and coordination. The main contributions of this paper are elaborated as follows:
\begin{itemize}
    \item We propose a novel network architecture that deploys MF-RIS on LEO satellites. MF-RIS is capable of extending the transmission range of LEO by reflecting, transmitting, and amplifying LEO signals. Moreover, MF-RIS can harvest partial signal as energy for compensating high energy consumption of LEO, especially when moving through the shadow areas where solar panels of LEOs cannot work.
    
    \item We aim at maximizing the long-term EE by determining MF-RIS configurations, including amplification, phase-shifts, and energy harvesting ratios as well as LEO transmit beamforming. To address the complex problem, a federated learning enhanced multi-agent deep deterministic policy gradient (FEMAD) scheme is designed. Multi-agent DDPG provides the optimal action policy from its interaction to environments, whereas federated learning enables cooperative information exchange among multi-agents.

    \item Numerical results have demonstrated that FEMAD can achieve the highest EE compared to other existing methods of centralized DRL and distributed MADDPG. The proposed architecture of LEO-MF-RIS achieves the highest EE among the scenarios of fixed/no energy harvesting in MF-RIS, conventional reflection-only RIS, and deployment without RISs/MF-RISs.
\end{itemize}

\section{System Model and Problem Formulation}

\begin{figure}[!t]
\includegraphics[width=3.5in]{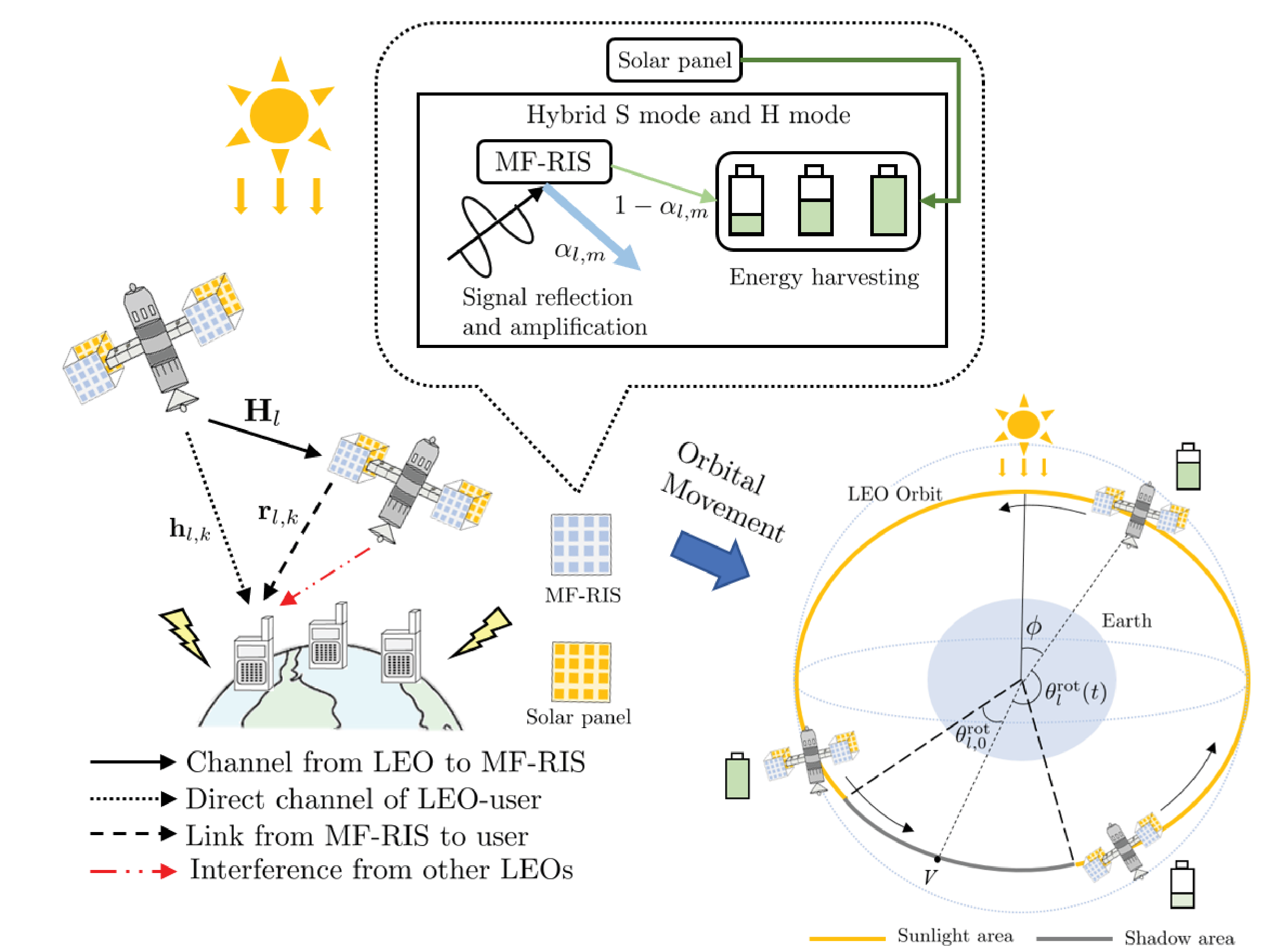}
\caption{The proposed architecture of LEO-MF-RIS. MF-RIS is equipped on LEO satellites, enabling hybrid signal and energy harvesting mode.} \label{Fig.1}
\end{figure}

	Fig. \ref{Fig.1} illustrates an MF-RIS assisted downlink LEO satellite communication network. We consider $L$ LEO satellites denoted by the set $\mathcal{L} = \{1, 2, \ldots, L\}$, with each equipped with $N$ transmit antennas, serving $K$ users indexed by the set of $ \mathcal{K} = \{1, 2, \ldots, K\}$. We assume a single receiving antenna for each user. An MF-RIS with $M$ elements indexed by the set of $\mathcal{M} = \{1, 2, \ldots, M\}$ is equipped on each LEO. Note that MF-RIS is in a two-dimensional array with $ M = M_h  \cdot M_v$ elements, where $M_h$ and $M_v$ indicate the respective numbers of elements in horizontal and vertical directions. The MF-RIS configuration on the $l$-th LEO can be defined as
\begin{align}
\mathbf{\Theta}_{l} = \operatorname{diag}\bigg( 
    \alpha_{l,1} \sqrt{\beta_{l,1}} e^{j\theta_{l,1}}, \ldots, \alpha_{l,M} \sqrt{\beta_{l,M}} e^{j\theta_{l,M}} 
\bigg),
\label{3}
\end{align}
where $ \theta_{l,m} \in [0, 2\pi) $ and $ \beta_{l,m} \in [0, \beta_{\text{max}}] $ denote the phase-shift and amplitude coefficients of MF-RIS, respectively. Note that $\beta_{\text{max}} > 1$ denotes the signal amplification, whereas $\beta_{\text{max}}  \leq 1$ indicates conventional RIS without amplification capability. Each element of the MF-RIS can operate in energy harvesting (EH) mode (H mode) and signal mode (S mode) by adjusting the EH coefficient $\alpha_{l,m}\in [0, 1]$. Note that $\alpha_{l,m} = 1$ implies that MF-RIS operates in S mode, whilst $\alpha_{l,m} = 0$ indicates that it functions in only H mode. MF-RIS can operate in a hybrid mode when $0< \alpha_{l,m} <1$, which is different from the existing work of \cite{6}. We define $s_{l,m}$ as the incident signal received by the $m$-th element of the MF-RIS on the $l$-th LEO satellite. Therefore, the signals harvested and reflected by the $m$-th MF-RIS element on the $l$-th LEO are modeled as $y_{l,m}^{h} = (1- \alpha_{l,m})s_{l,m}$ and $ y_{l,m}^{r} =\alpha_{l,m}\sqrt{\beta_{l,m}} e^{j\theta_{l,m}} s_{l,m}$, respectively.

\subsection{Channel Model}
	
	We consider the Rician fading channel model \cite{13} between the the $l$-th LEO and MF-RIS as $\mathbf{H}_{l} \in \mathbb{C}^{M \times N}$, i.e., 
\begin{align}
	\mathbf{H}_l = \sqrt{h_0 d_{sr}^{-k_0}}
\left( \sqrt{\frac{\beta_0}{\beta_0 + 1}} \mathbf{H}_{\text{LoS}} + \sqrt{\frac{1}{\beta_0 + 1}} \mathbf{H}_{\text{NLoS}} \right),\
\label{4}
\end{align}
where $h_0$ is the pathloss at the reference distance of 1 meter , $d_{sr}$ is the distance, and $k_0$ is the corresponding pathloss exponent. $\beta_0$ is the Rician factor, adjusting the portion of LoS path  $\mathbf{H}_{\text{LoS}}$ and non-LoS (NLoS) component $\mathbf{H}_{\text{NLoS}}$. The LoS component $ \mathbf{H}_{\text{LoS}} $ is expressed as \cite{14}
\begingroup
\allowdisplaybreaks
\begin{align}
&\mathbf{H}_{\text{LoS}} 
	= 
\begin{bmatrix}
1,e^{-j\frac{2\pi}{\lambda}d\sin\varphi_r\sin\vartheta_r}, \cdots,e^{-j\frac{2\pi}{\lambda}(M_z-1)d\sin\varphi_r\sin\vartheta_r}
\end{bmatrix}^T \notag \\
	&\otimes
\begin{bmatrix}
1,e^{-j\frac{2\pi}{\lambda}d\sin\varphi_r\cos\vartheta_r}, \cdots,e^{-j\frac{2\pi}{\lambda}(M_y-1)d\sin\varphi_r\cos\vartheta_r}
\end{bmatrix}^T \notag \\
	&\otimes
\begin{bmatrix}
1,e^{-j\frac{2\pi}{\lambda}d\sin\varphi_t\cos\vartheta_t}, \cdots,e^{-j\frac{2\pi}{\lambda}(N-1)d\sin\varphi_t\cos\vartheta_t}
\end{bmatrix}^T,
\label{5}
\end{align}
\endgroup
where $\otimes$ denotes the Kronecker product and $T$ is transpose operation. Notation of $\lambda$ indicates the wavelength of the operating frequency, and $d$ denotes the element separation of MF-RIS. In \eqref{5}, $\varphi_r$, $\vartheta_r$, $\varphi_t$, and $\vartheta_t$ represent the vertical/horizontal angle-of-arrivals, and the vertical/horizontal angle-of-departures, respectively. Note that $\mathbf{H}_{\text{NLoS}}$ follows independent and identically distributed Rayleigh fading. The channel vectors from LEO $l$ to user $k$ and from the MF-RIS $l$ to user $k$, denoted by $\mathbf{h}_{l,k} \in \mathbb{C}^{N \times1 }$ and $\mathbf{r}_{l,k} \in \mathbb{C}^{M \times 1 }$, are defined by following $\mathbf{H}_{l} $ but in a vector form, where the LoS components $ \mathbf{h}_{\text{LoS}}$ and $\mathbf{r}_{\text{LoS}}$ are expressed as
\begin{align}
	& \mathbf{h}_{\text{LoS}} = \left[
1, e^{-j\frac{2\pi}{\lambda}d\sin\varphi_t\sin\vartheta_t}, \cdots, e^{-j\frac{2\pi}{\lambda}(N-1)d\sin\varphi_t\sin\vartheta_t}
\right]^T, \\
	& \mathbf{r}_{\text{LoS}}= \left[
1, e^{-j\frac{2\pi}{\lambda}d\sin\varphi_t\sin\vartheta_t}, \cdots, e^{-j\frac{2\pi}{\lambda}(M-1)d\sin\varphi_t\sin\vartheta_t}
\right]^T.
\end{align}
For simplicity, we neglect the remaining parameters of $\mathbf{h}_{l,k}$ and $\mathbf{r}_{l,k}$ due to similar definitions in \eqref{4} and \eqref{5}. Accordingly, the combined channel between LEO-MF-RIS $l$ and user 
$k$ can be expressed as $\mathbf{g}_{l,k} \in \mathbb{C}^{1 \times N }$
\begin{align}
	\mathbf{g}_{l,k} =\mathbf{h}_{l,k}^H  + \mathbf{r}_{l,k}^H \mathbf{\Theta}_{l} \mathbf{H}_{l},
\label{7}
\end{align}
where $H$ indicate Hermitian operation. The transmitted signal and the active beamforming vector of the $k$-th user served by the $l$-th LEO are defined as $x_{l,k}$ and $ \mathbf{w}_{l,k} \in \mathbb{C}^{N \times 1}$, respectively. Consequently, the received signal of the $k$-th user served by the $l$-th LEO is given by
\begin{align} \label{r_signal}
y_{l,k} &= \mathbf{g}_{l,k} \mathbf{w}_{l,k} x_{l,k} 
+ \mathbf{g}_{l,k} \sum_{k' \in \mathcal{K} \backslash k} \mathbf{w}_{l,k'} x_{l,k'} \notag \\
&\quad + \sum_{l' \in \mathcal{L}  \backslash l} \sum_{k' \in \mathcal{K} \backslash k} \mathbf{g}_{l',k'} \mathbf{w}_{l',k'} x_{l',k'}+n_{l,k},
\end{align}
where $ n_{l,k} \sim \mathcal{CN}(0, \sigma_{l,k}^2) $ denotes complex additive white Gaussian noise (AWGN) with power $ \sigma_{l,k}^2$. In \eqref{r_signal}, the first term represents the desired signal for user $k$ from the $l$-th LEO. The second term denotes the intra-LEO interference, whilst the third term represents inter-LEO interference. According to \eqref{r_signal}, the corresponding signal-to-interference-plus-noise ratio (SINR) is given by
\begingroup
\allowdisplaybreaks
\begin{align} \label{sinr}
\gamma_{l,k} = 
\frac{\left| \mathbf{g}_{l,k} \mathbf{w}_{l,k} \right|^2}{
    \sum\limits_{k' \in \mathcal{K} \backslash k} \left| \mathbf{g}_{l,k} \mathbf{w}_{l,k'} \right|^2 
    + \sum\limits_{l' \in \mathcal{L} \backslash l} 
    \sum\limits_{k' \in \mathcal{K} \backslash k} \left| \mathbf{g}_{l',k} \mathbf{w}_{l',k'} \right|^2 
    + \sigma_{l,k}^2}.
\end{align}
\endgroup
Based on \eqref{sinr}, the ergodic rate of LEO $l$ and user $k$ can be obtained as
\begin{align}
R_{l,k} = \log_2 (1 + \gamma_{l,k}).
\end{align}
We also note that time step $t$ is omitted in some content of this article for simplicity, i.e., $R_{l,k}(t)$ is written as $R_{l,k}$, as the movement of LEO satellites is predetermined by their orbital paths.

\subsection{Power Dissipation Model} 
	We define the EH coefficient matrix for the $ m $-th element of the $l$-th MF-RIS on LEO satellite as
\begin{align}
\mathbf{T}_{l,m} = \operatorname{diag}\left([ 0, \cdots, 0, 1 - \alpha_{l,m}, 0, \cdots, 0 ]\right). \label{eq:11}
\end{align}
Accordingly, the RF power received at the $ m $-th element of the $l$-th MF-RIS is acquired as
\begin{align}
P_{l,m}^{\text{RF}} =  \left\lVert \mathbf{T}_{l,m} \left( \mathbf{H}_{l} \sum_{l\in \mathcal{L}} \sum_{k\in \mathcal{K}} \mathbf{w}_{l,k} + \mathbf{n}_{m} \right) \right\rVert^2 ,
\end{align}
where $ \mathbf{n}_{m} \sim \mathcal{CN}(0, \sigma_{m}^2 \mathbf{I}_M) $ denotes the amplified noise introduced by the MF-RIS with its power $\sigma_m^2$. In order to capture the dynamics of the RF energy conversion efficiency for different input power levels, a non-linear energy harvesting model is adopted \cite{15}. Accordingly, the total power harvested from the $ m $-th element of the $l$-th MF-RIS on LEO is given by
\begin{align}
P_{l,m}^h = \frac{\Upsilon_{l,m} - Z \Omega}{1 - \Omega},\label{12}
\end{align}
where $ \Upsilon_{l,m} = \frac{Z}{1 + e^{-a (P_{l,m}^{\text{RF}} - q)}} $ is a logistic function with respect to the received RF power $ P_{l,m}^{\text{RF}} $, and $ Z \geq 0 $ is a constant determining the maximum harvested power. The constant $\Omega = \frac{1}{1 + e^{a q}}$ is introduced to ensure a zero-input/zero-output response for H mode, with constants $ a > 0 $ and $ q > 0 $ capturing the joint effects of circuit sensitivity limitations and current leakage. Moreover, the power for controlling MF-RIS mainly comes from the total number of PIN diodes required \cite{thzris}. The total number of required PIN diodes can be obtained as $\log_2 L_{\alpha} + \log_2 L_{\beta} + 2 \log_2 L_{\theta}$, where $L_{\alpha}$, $L_{\beta}$, and $L_{\theta}$ indicate the quantization levels controlled by PIN diodes for the EH coefficient, amplitude, and phase-shifts, respectively. Then, the power consumption of the MF-RIS can be obtained as
\begin{align} \label{Rcon}
P_{l,R}^{\text{cons}} \!=\! \frac{1}{2} \left[ \log_2 L_{\alpha} \!+\! \log_2 L_{\beta} \!+\! 2 \log_2 L_{\theta} \right] M P_{\text{pin}} \!+\! P_C \!+\! \xi P_{l,O},
\end{align}
where $P_C$ is the power consumed by the RF-to-DC power conversion circuit, $P_{l,O}$ is the output power of the MF-RIS, and $\xi$ is the inverse of the amplifier efficiency. In \eqref{Rcon}, the output power of the MF-RIS can be expressed as
\begin{equation}
P_{l,O} = \sum_{k\in \mathcal{K}} \left\| \mathbf{\Theta}_{l} \mathbf{H}_{l} \mathbf{w}_{l,k} \right\|^2 + \sum_{m\in \mathcal{M}}\sigma_m^2 \left\| \mathbf{\Theta}_{l} \right\|_F^2,
\end{equation}
where $ \|\cdot\|_F $ is the Frobenius norm.


We partition it into multiple time slots, with the interval of each time slot defined as $\Delta$. In time $t$, the energy harvested by satellite solar panels can be expressed
as \cite{16}
\begin{equation}
	E_{l}^{\text{sol}}(t) = \int_t^{t+\Delta} \eta_s \psi B \sqrt{1 - \cos^2\phi \cos^2\theta_l^{\text{rot}}(t)} \, {\rm d}t,
\end{equation}
where $\eta_s$ is the efficiency of energy conversion, $\psi$ is the light intensity, and $B$ is the size of the solar panels. The notation of $\phi$ indicates the angle between the satellite orbital plane and the sunlight, and $\theta_l^{\text{rot}}(t)$ defines the satellite rotation angle from the midpoint of the shaded orbit $V$ at time $t$. Fig. \ref{Fig.1} illustrates an example of the LEO satellite's energy consumption when moving over sunlight and shadow areas. In sunlight area, the battery is charged by harvesting the solar energy from both solar panels and signal energy from the MF-RIS. However, in the shadow region where solar energy is unavailable, the LEO relies primarily on the remained battery capacity and additional energy harvested from the MF-RIS. We define the half-angle of the shadow area that the LEO satellite travels as \cite{16}
\begin{align}
\theta_{l,0}^{\text{rot}} = 
\begin{cases} 
0, \quad \text{ if } \phi > \sin^{-1} \left( \frac{R_e}{R_e + h_{l,s}} \right) ,\\
\sin^{-1} \left( \frac{R_e^2 \cos^2\phi - \left(2 R_e h_{l,s} + h_{l,s}^2 \right) \sin^2\phi}{(R_e + h_{l,s}) \cos \phi} \right), & \text{otherwise}, 
\end{cases}
\end{align}
where $R_e$ is the radius of the Earth and $h_{l,s}$ is the height of the $l$-th LEO satellite. Note that the range of  $\theta_{l,0}^{\text{rot}}$ is $(- \pi, \pi)$. When the satellite is in the sunlight area, i.e., $|\theta_l^{\text{rot}}(t)| \geq \theta_{l,0}^{\text{rot}}$, the remaining time it takes to move from its current position $\theta_l^{\text{rot}}(t)$ to the shadow area can be expressed as
\begin{equation}
	T_{l}^{\text{sun}}(t) = 
\begin{cases} 
\frac{2\pi - \theta_{l,0}^{\text{rot}} - \theta_l^{\text{rot}}(t)}{\dot{\Omega}}, & \text{ if }\theta_l^{\text{rot}}(t) \in [0, \pi), \\
\frac{-\theta_{l,0}^{\text{rot}} - \theta_l^{\text{rot}}(t)}{\dot{\Omega}}, & \text{ if }\theta_l^{\text{rot}}(t) \in [-\pi, 0) ,
\end{cases}
\end{equation}
where $\dot{\Omega}$ is the Earth rotation rate. Moreover, the duration of LEO $l$ moving from the current location in shadow area to the sunlight area can be expressed as
\begin{align}
	T_{l}^{\text{shd}}(t) = \frac{\theta_{l,0}^{\text{rot}} - \theta_l^{\text{rot}}(t)}{\dot{\Omega}}.
\end{align}
According to \cite{16, 17}, the charging solar power of the LEO satellite battery at time $t$ can be expressed as
\begin{equation}
	P_{l,\text{in}}^b(t) = \frac{E_{l}^{\text{sun}}(t)}{T_{l}^{\text{sun}}(t)}.
\end{equation}
When the LEO satellite is in the sunlight area, the remaining battery energy on LEO $l$ at time  $t$ can be expressed as
\begingroup
\allowdisplaybreaks
\begin{align}
& E_{l}(t) = 
	\min\left( E^b,  E_{l}(t-1) + \left[ P_{l,\text{in}}^b(t)\cdot \mathbbm{1}(\mathcal{A}=\text{sun}) \right.\right. \notag \\
	& \left.\left. +\sum_{m \in \mathcal{M}} P_{l,m}^h - P_{l,R}^{\text{cons}} - (P_{l}^{\text{tr}} + P_{l}^{\text{cons}}) \right] \cdot T_{l}^{\mathcal{A}}(t-1) \right), \label{19}
\end{align}
\endgroup
where $\mathcal{A}\in \{ \text{sun}, \text{shd}\}$ indicates whether the current LEO location is in sunlight or shadow area, and $\mathbbm{1}(\cdot)$ indicates the event occurrence. In \eqref{19}, $E^b$ denotes the LEO satellite battery capacity, $P^{\text{tr}}_{l} = \sum_{k \in \mathcal{K}} \|\mathbf{w}_{l,k}\|^2$ indicates the transmit power of LEO $l$, and $P_{l}^{\text{cons}}$ represents the regular operational circuit power. The total energy consumption can be expressed as
\begin{equation}
E_{l}^{\text{tot}}(t) = \left( P_{l,R}^{\text{cons}} + P_{l}^{\text{tr}} + P_{l}^{\text{cons}}
- \sum_{m \in \mathcal{M}}P_{l,m}^h \right) \cdot T_{l}(t),
\end{equation}
where $T_{l}(t) = T_{l}^{\text{sun}}(t) + T_{l}^{\text{shd}}(t)$ denotes the total time it takes for the LEO satellite to complete one full orbit.

\subsection{Problem Formulation}
The objective is to maximize the long-term EE while guaranteeing the constraints of user rate requirement, MF-RIS configuration and LEO power limitation, which is formulated as follows:
\begingroup
\allowdisplaybreaks
\begin{subequations} \label{prob}
	\begin{align}
& \max_{\substack{\theta_{l,m}, \alpha_{l,m}, \\ \beta_{l,m},\mathbf{w} _{l,k}}}  \ \lim_{T \to \infty} \frac{1}{T} \sum_{t=0}^{T} \sum_{l\in 
\mathcal{L}} \sum_{k\in \mathcal{K}} \frac{R_{l,k}(t)}{E_{l}^{\text{tot}}(t)} \label{prob_25} \\
	& \text{s.t.} \quad \mathbf{\Theta}_{l} \in \mathcal{R}_{\mathbf{\Theta}}, \qquad\quad\quad\ \ \forall l \in \mathcal{L}, \label{25b}\\
	& \qquad R_{l,k} \geq R^{\text{min}}_{l,k}, \qquad\qquad \forall l \in \mathcal{L}, \forall k \in \mathcal{K}, \label{25c} \\
	& \qquad P^{\text{cons}}_{l,R} \leq \sum_{m \in \mathcal{M}} P^{h}_{l,m}, \quad\ \forall l \in \mathcal{L}, \label{25d} \\
	& \qquad P^{\text{tr}}_l + P^{\text{cons}}_l \leq P^{b}_{l,o}, \quad \ \, \forall l \in \mathcal{L}, \label{25e} \\
	& \qquad E_l(t) \geq 0, \qquad\qquad\quad \forall l \in \mathcal{L}. \label{25f}
	\end{align}
\end{subequations}
\endgroup
Constraints \eqref{25b} defines the constraint set of MF-RIS coefficients as $\mathcal{R}_{\mathbf{\Theta}}$, including $\alpha_{l,m} \in [0,1]$, $\beta_{l,m} \in [0, \beta_{\text{max}}]$, and $\theta_{l,m} \in [0, 2\pi)$. Constraint \eqref{25b} guarantees the minimum rate requirement of each serving users as $R_{l,k}^{\text{min}}$. Constraint \eqref{25c} ensures the self-sustainability of MF-RIS, i.e., power consumption of the MF-RIS cannot exceed its harvested power. Constraint \eqref{25d} guarantees that the total LEO transmit power and circuit operational power should be smaller than $P^b_{l,o}$. Constraint \eqref{25e} ensures that the remaining battery energy must be greater than $0$. Due to the inherent non-convexity and non-linearity of problem \eqref{prob_25}, as well as the complexity of long-term EE optimization, it presents a significant challenge to solve this problem. To address these difficulties, we propose a DRL-based scheme, which is detailed in the following section.

\section{Proposed Federated Learning Multi-Agent Deep Deterministic Policy Gradient (FEMAD) Scheme}


We have conceived a FEMAD scheme for solving problem \eqref{prob}, as depicted in Fig. \ref{Fig3}. FEMAD consists of multi-agent DDPG architecture by allowing each LEO as an agent determining its own policy with a plethora of continuous variables. Furthermore, federated learning is leveraged in the multi-agent system for cooperative hidden information exchange.

\subsection{Multi-Agent DDPG Algorithm}

	We consider a typical DRL framework consisting of a state set $\mathcal{S}$, an action set $\mathcal{A}$, and a reward set $\mathcal{R}$. In this framework, an agent representing one LEO-MF-RIS will iteratively perform its action in a dynamic environment to receive the corresponding reward while updating its current state. Since optimizing MF-RIS in LEO networks involves multiple observation and configuration parameters, the state-action space becomes compellingly large. Traditional DRL methods would likely be inefficient, facing issues such as learning instability and slow convergence. Additionally, traditional deep Q networks (DQN) are unsuitable for this problem, as they could result in prohibitively high complexity due to the high-dimensional discrete variables when quantizing continuous parameters. For these reasons, we adopt the MADDPG architecture utilizing a two-layered actor-critic network to effectively address these issues. Accordingly, we define the state, action, and the corresponding reward as follows:
 \begin{itemize}
    \item \textbf{State:} The total state space is defined as a set of individual agent state $\mathcal{S}(t) = \{\mathcal{S}_1(t), \mathcal{S}_2(t), \ldots, \mathcal{S}_L(t)\}$. Each agent state $\mathcal{S}_l(t)$ is composed of a sequence of states over time, denoted as $\mathcal{S}_l(t) = \{s_l(1), s_l(2), \cdots, s_l(T)\}$, where $s_l(t) = \{ \mathbf{g}_{l,k}(t) |  \forall t ,\forall k \in \mathcal{K} \}$, with $\mathbf{g}_{l,k}(t)$ defined as the combined MF-RIS channel in \eqref{7} at time $t$.

    \item \textbf{Action:} The action space $ \mathcal{A}(t) $ is defined as a set of individual agent action $ \mathcal{A}(t) = \{\mathcal{A}_1(t), \mathcal{A}_2(t), \ldots, \mathcal{A}_L(t)\} $. Each agent action set $ \mathcal{A}_l(t) $ consists of a sequence of actions over time, denoted as $ \mathcal{A}_l(t) = \{a_l(1), a_l(2), \cdots, a_l(T)\} $. Note that action $a_l(t)$ includes all parameters to be determined in problem \eqref{prob_25} as $ a_l(t) = \{\theta_{l,m}(t), \alpha_{l,m}(t), \beta_{l,m}(t), \mathbf{w}_{l,k}(t) | \forall t, \forall m\in \mathcal{M}, \forall k\in \mathcal{K}\} $.
    
    \item \textbf{Reward:} 
The reward space set is defined as $\mathcal{R}(t) = \{\mathcal{R}_1(t), \mathcal{R}_2(t), \ldots, \mathcal{R}_L(t)\}$, with each agent reward set as $\mathcal{R}_l(t) = \{r_l(1), r_l(2), \cdots, r_l(T)\}$. The obtained reward of each agent is designed as the penalized EE as
\begin{align}
r_l(t) = \frac{\sum_{k\in \mathcal{K}} R_{l,k}(t)}{E_{l}^{\text{tot}}(t)} - \sum_{i=1}^{4} \rho_{l,i} C_{l,i},
\end{align}
where $\rho_{l,i}, \forall i\in \{1,2,3,4\}, \forall l\in \mathcal{L}$ indicates the weights of each penalty $C_{l,i}$. The penalty terms $C_{l,i}$ corresponding to constraints of \eqref{25c}--\eqref{25f} are defined as $ C_{l,1} = \sum_{k\in\mathcal{K}} R_{l,k}^{\text{min}} - R_{l,k} $, $ C_{l,2} = P_{l,R}^{\text{cons}}-\sum_{m \in \mathcal{M}} P^{h}_{l,m}$, $ C_{l,3} = P^{\text{tr}}_l + P^{\text{cons}}_l - P^{b}_{l,o} $, and $ C_{l,4} = - E_l(t)$.
\end{itemize}

\begin{figure}[t] 
\centering
\includegraphics[width=3.5in]{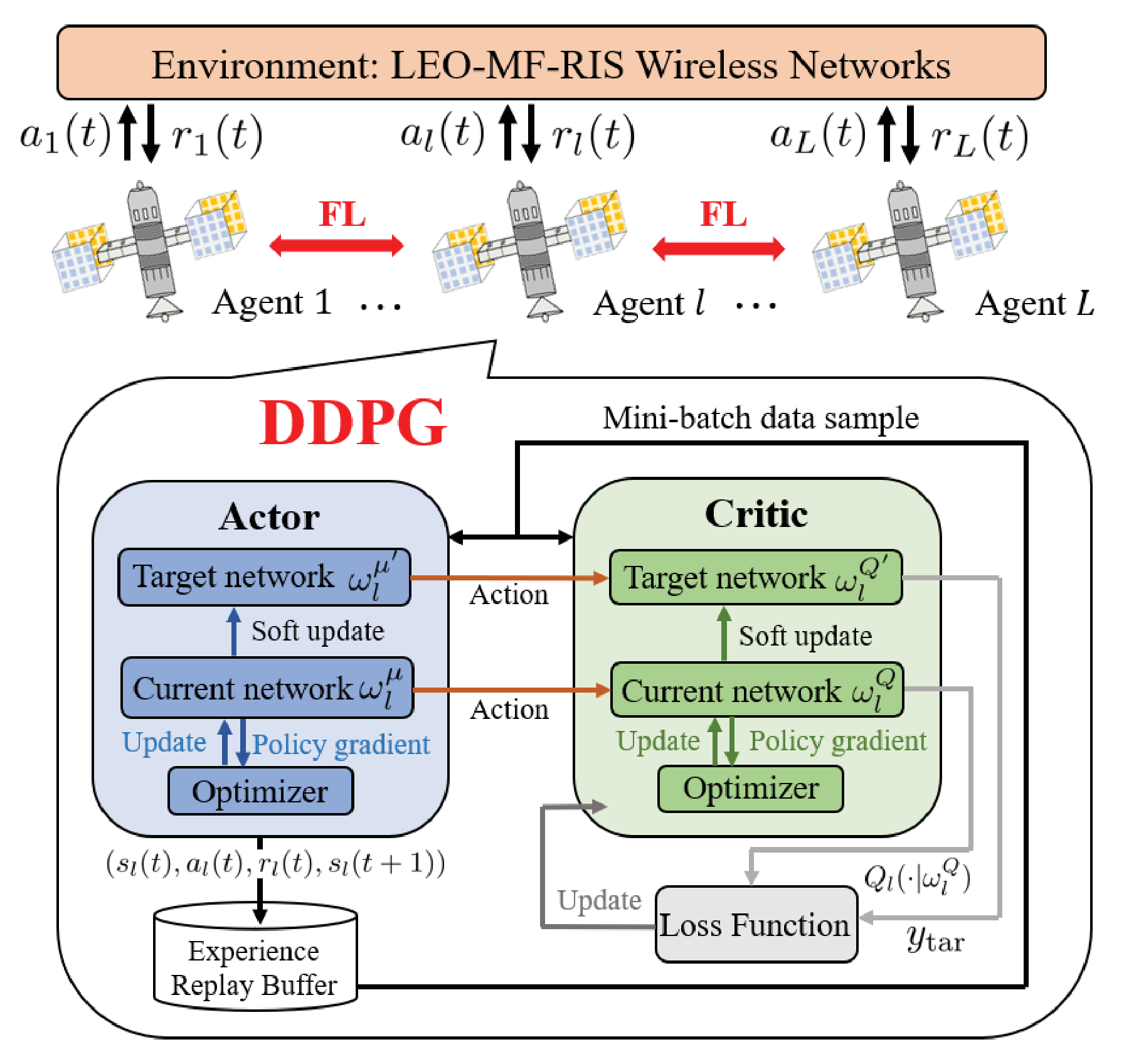}
\caption{The proposed FEMAD architecture for LEO-MF-RIS networks.} 
\label{Fig3}
\end{figure}

As shown in Fig. \ref{Fig3}, MADDPG includes both the current network and the target network, with each consisting of an actor and a critic sub-network. Note that network here indicates the deep neural network of DDPG. Here, $ \omega^\mu_l $ and $ \omega^Q_l $ represent the actor/critic model weights of the current network, whilst $ \omega^{\mu'}_l $ and $ \omega^{Q'}_l $ represent the actor/critic model weights of the target network. The current network determines the action of the $l$-th LEO-MF-RIS as $ a_l(t) = \mu_l(s_l(t)|\omega^\mu_l) + N_G $, where $ \mu_l(s_l(t)|\omega^\mu_l) $ is the output of the actor network given the input state $s_l(t)$. Notation of $N_G$ is defined as Gaussian noise, which enables exploration for potential new actions in the environment. Moreover, the critic network evaluates the Q-value by taking the state and action as inputs, providing an assessment of the expected return as
\begin{align}
    &Q_l\left(s_l(t), a_l(t) | \omega^Q_l \right) = r_l(t)+ \gamma_{l} \cdot  \nonumber \\
    & \qquad Q_l \left(s_l(t+1), a_l(t+1), \ldots, a_{l'\neq l}(t+1) \mid \omega^{Q'}_{l} \right),
\end{align}
where the discount factor $ \gamma \in [0, 1] $ indicates the importance of future rewards. Note that $a_{l'\neq l}$ means the actions taken by other LEO agents contributing to the common environment. We define the memory replay buffer $\mathcal{D}$ with a size of $|\mathcal{D}|$. The tuple of $(s_l(t), a_l(t), r_l(t), s_l(t+1))$ will be stored into the memory for further learning. During training of MADDPG, we randomly select $X$ samples from $\mathcal{D}$ for training, with the loss function given by
\begin{equation}
    L(\omega^{Q}_{l}) = \frac{1}{X} \sum_{t=1}^{X} \left(y_{\text{tar}} - Q_l \left(s_l(t), a_1(t), \ldots, a_L(t) \mid \omega^{Q}_{l} \right) \right)^2,
\end{equation}
where $y_{\text{tar}} = r_l(t) + \gamma \cdot {Q}_l^{\prime} \left(s_l(t+1), a'_1, \ldots, a'_L \mid \omega^{Q'}_{l} \right)$. Note that $X \leq |\mathcal{D}|$. The stochastic gradient descent is employed to update the model weights of the current network, with the gradient computed as $\nabla_{\omega^{\mu}_{l}} L(\omega^{Q}_{l})  \approx \frac{1}{X} \sum_{t=1}^{X} \nabla_{a_{l}(t)} Q_l \left(s_l(t), a_1(t), \ldots, a_L(t) \mid \omega^{Q}_{l} \right) \cdot \nabla_{\omega^{\mu}_{l}} \mu_{l} \left(s_l(t) \mid \omega^{\mu}_{l} \right)$. Furthermore, the target network will periodically update the model weights from the current network based on the soft update \cite{17} for both actor-critic sub-networks, which is represented by $\omega_{l}^{\mu'} \leftarrow \tau_{\mu} \omega^{\mu}_{l} + (1 - \tau_{\mu}) \omega_{l}^{\mu'}$ and $\omega_{l}^{ Q'} \leftarrow \tau_{Q} \omega^{Q}_{l} + (1 - \tau_{Q}) \omega_{l}^{ Q'}$,
where $ 0\leq \tau_{\mu} $ , $ \tau_{Q}\leq 1 $ are positive constants indicating the significance of actor parameters of the target and current networks, respectively.


\subsection{Enhancing MADDPG with Federated Learning }

In the traditional MADDPG framework, the lack of information exchange between agents might result in suboptimal solution, especially in highly complex multi-agent environments that require collaborative decision-making. To address this limitation, we propose an enhanced approach by incorporating federated learning (FL) into the MADDPG architecture as the proposed FEMAD scheme. In FEMAD, the model weight of the target critic network from the local LEO-MF-RIS agent  are exchanged. We consider the model exchange is performed by the selected LEO edge with the best channel quality. Furthermore, we conduct communication-efficient FL, i.e,, partial nodes of the model are exchanged for preventing high-latency under distant LEO communications \cite{17,my3}. The LEO edge server will aggregate these local models to exchange the implicit information of other LEOs' policies and potential environmental interference, which can be expressed as
\begin{equation}
	\omega^{\text{global}} = \frac{1}{L} \sum_{l\in \mathcal{L}} \xi_l  \cdot \omega^{\text{local}}_l,
\end{equation}
where $\xi_l \in [0, 1]$ represents the importance ratio of each training model and $\sum_{l\in \mathcal{L}} \xi_l = 1$. Note that $\xi_l = 1$ is considered, indicating equivalent importance of individual hidden information of LEO-MF-RIS. After completing the model aggregation, the LEO edge will broadcast the global model to each local LEO agent for further training.

\section{Simulation result}

\begin{table}[!t]
\scriptsize
\caption{Simulation parameters} \label{param}
\centering
\begin{tabular}{|p{2.1cm}|p{5cm}|} 
\hline
\textbf{Parameter} & \textbf{Value} \\ \hline
Communication parameters & 
$N=16$, $M=16$, $L=10$, $K=20$, $h_0 = -20$ dB, $\kappa_0 = 2.2$, $\beta_0  = 3$ dB, $\sigma_{l,k}^2 = \sigma_m^2 = -70$ dBm \cite{6} \\ \hline
MF-RIS power consumption
parameters &
$\xi = 1.1$, $P_{\text{pin}} = 0.33$ mW, $P_{l}^{\text{cons}} = 90$ W, 
$P_C = 10$ W, $Z = 24$ mW,  
$a = 150$, $q = 0.014$ \cite{6} \\ \hline
Other LEO-related parameters &  $L_{\alpha}=2$, $L_{\beta}=10$, $L_{\theta}=8$, $E^b = 9 \times 10^4$ J, $\eta_s=0.19$, $\psi=500$ W/$\text{m}^2$, $B=4$ $\text{m}^2$, $R_e=6378$ km, $h_{l,s}=1000$ km, $\dot{\Omega}=7.29 \times 10^{-5}$ rad/s \cite{16} \\ \hline
\end{tabular}
\label{table:4x2_example}
\end{table}

\begin{figure}[t]
\centering
\includegraphics[width=3.2in]{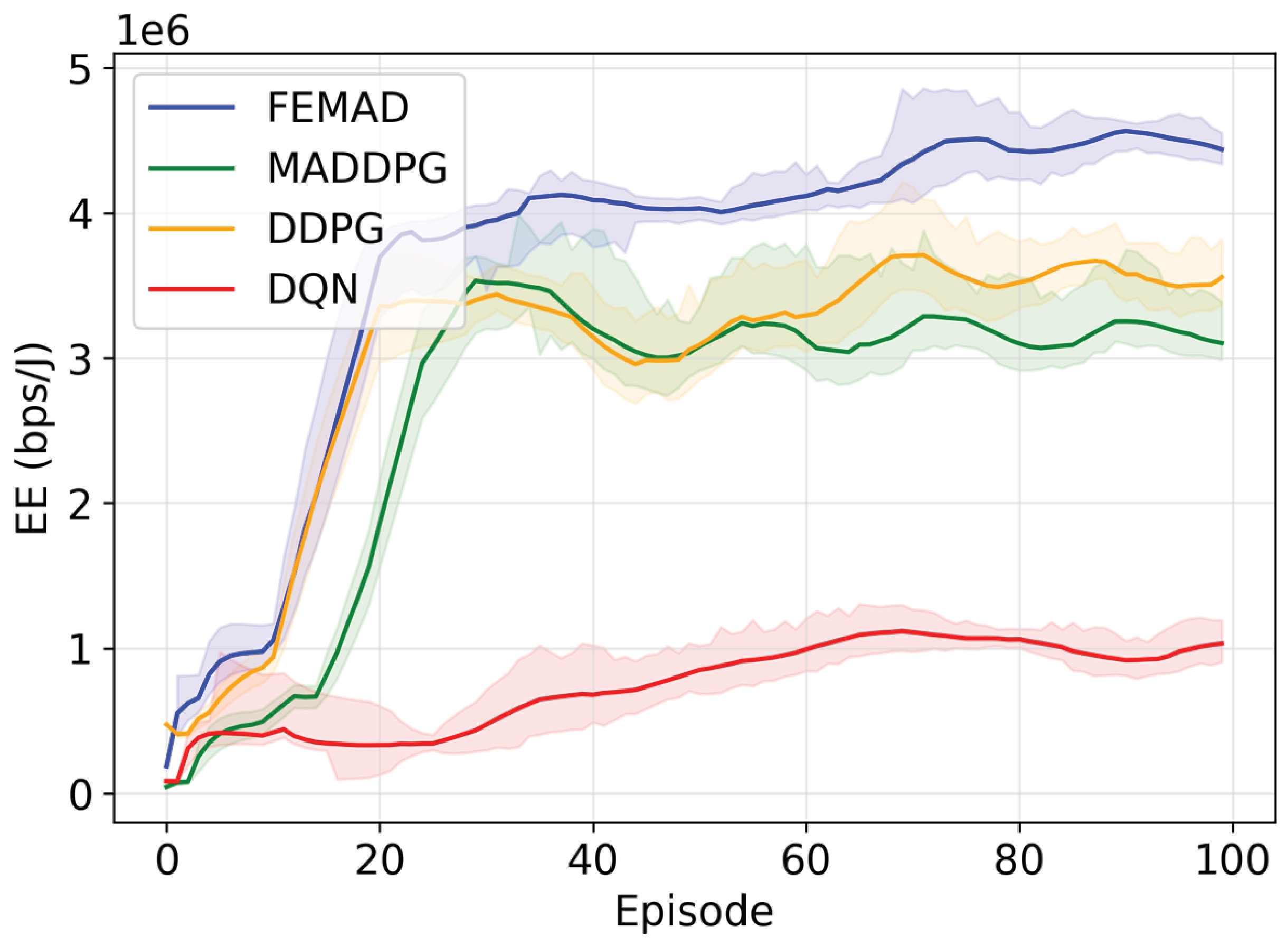}
\caption{Convergence of FEMAD and existing DRL methods.} \label{Fig.4}
\end{figure}

\begin{figure}[t]
\centering
\includegraphics[width=3.2in]{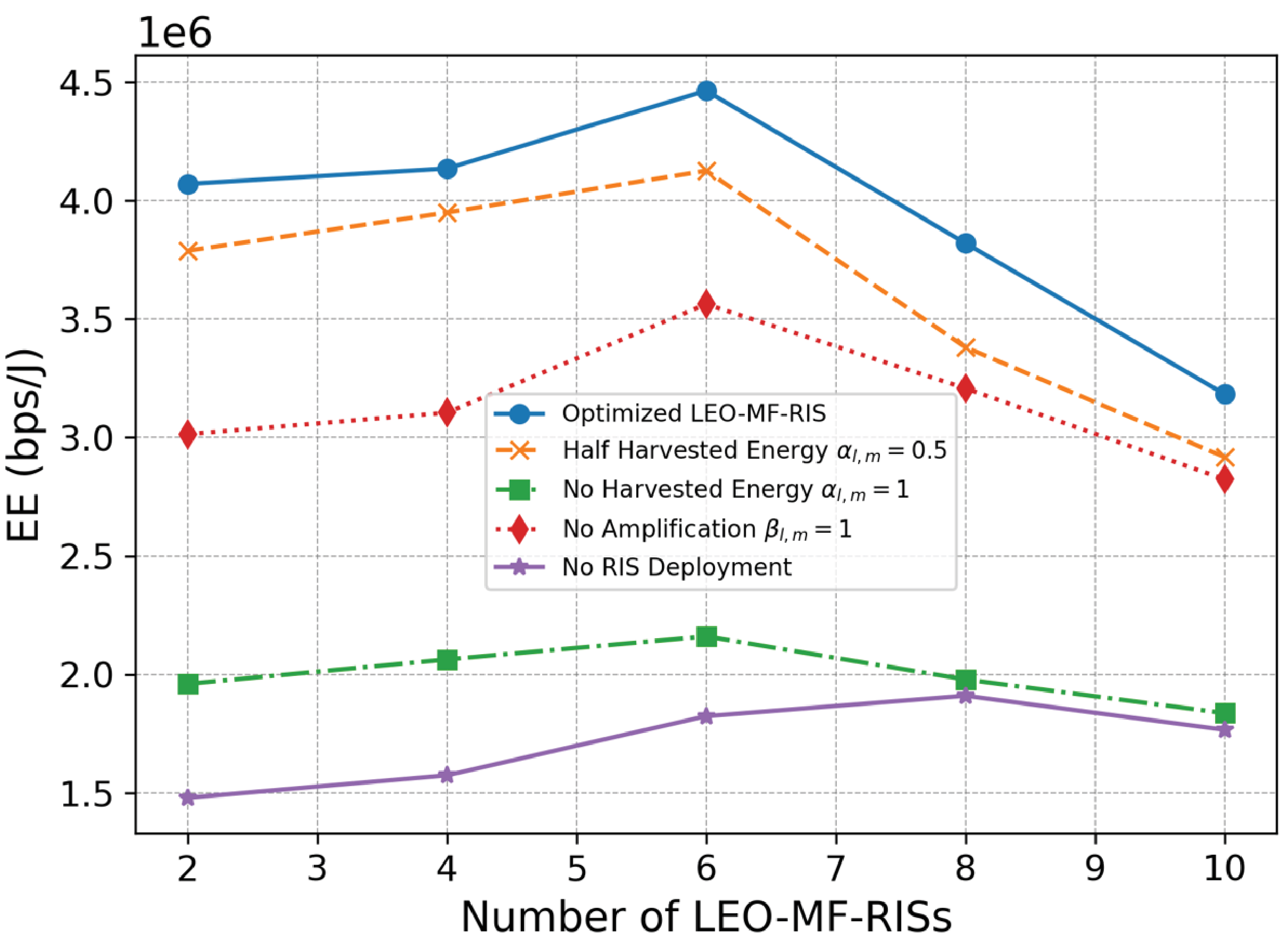}
\caption{EE with different MF-RIS configurations, i.e., no amplification, fixed/no energy harvesting, and no-RIS versus different numbers of LEOs.} \label{Fig.5}
\end{figure}

\begin{figure}[t]
\centering
\includegraphics[width=3.2in]{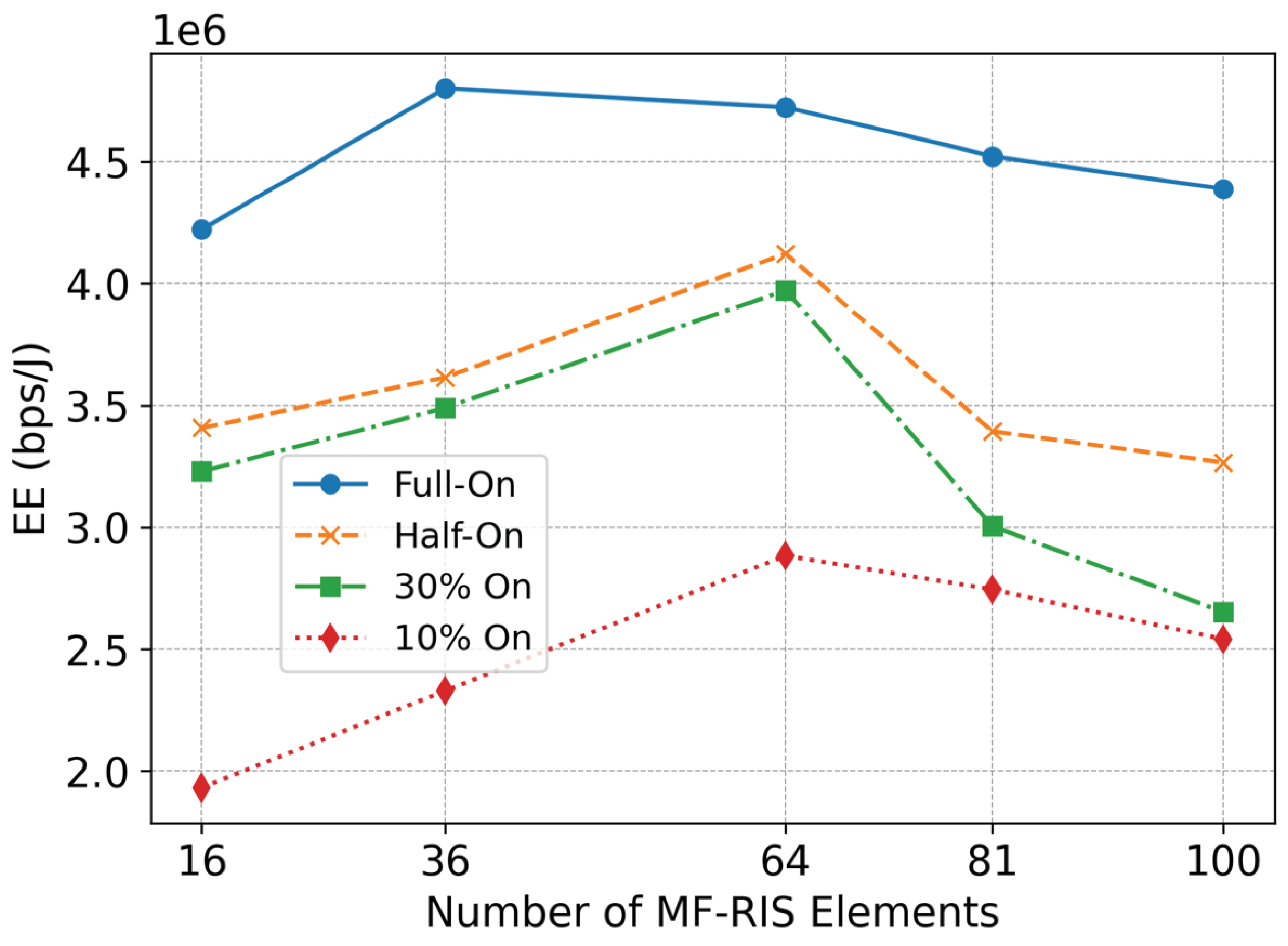}
\caption{EE versus different numbers of MF-RIS elements. A portion of elements are switched on, i.e., full-on, half-on, $30\%$ and $10\%$ of elements.} \label{Fig.6}
\end{figure}

\begin{figure}[t]
\centering
\includegraphics[width=3.2in]{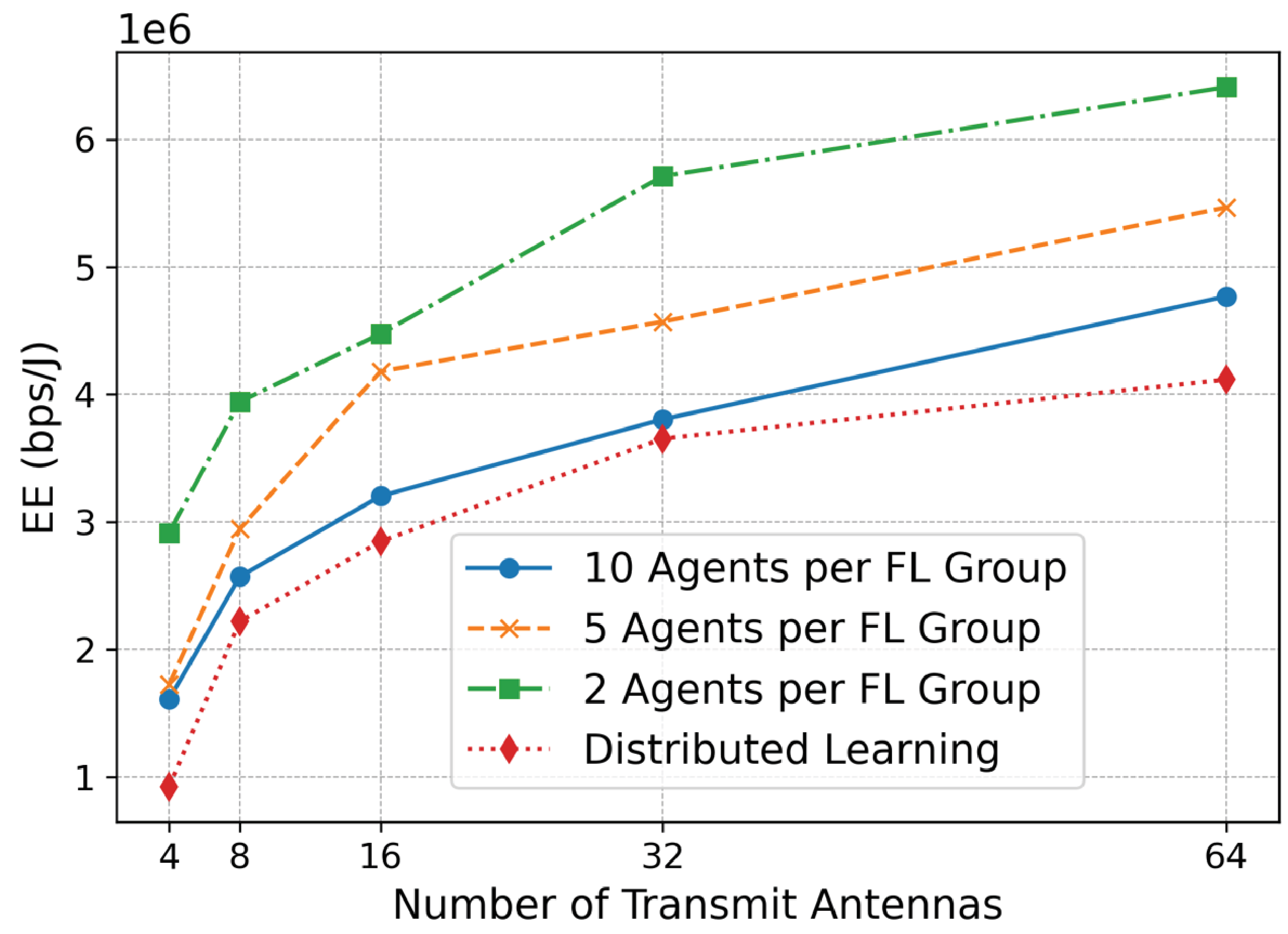}
\caption{EE performance of FEMAD with different numbers of LEOs in an FL group and of LEO transmit antennas.} \label{Fig.7}
\end{figure}


	In simulations, we construct the Globalstar LEO satellite system \cite{18} to evaluate the proposed LEO-MF-RIS performance. The pertinent parameters are listed in Table \ref{param}. As for FEMAD scheme, the learning rates of actor/critic networks are set to $0.001$ and $0.0005$, respectively. The discount factor is set to $\gamma = 0.99$, whilst soft update parameters are set to $\tau_{\mu}=\tau_{Q} = 0.005$. The experience replay buffer $\mathcal{D}$ can store up to $10^5$ samples. The mini-batch mechanism is adopted during training, with a batch size of $64$.

Fig. \ref{Fig.4} presents the convergence behavior of the proposed FEMAD algorithm compared to the existing DRL methods. The results have demonstrated that FEMAD achieves superior EE with faster convergence. We can observe that DDPG surpasses MADDPG by enabling centralized learning for search the global optimum solution. However, without model  exchange carrying useful hidden information from other LEO agents, MADDPG has around a $10\%$ degradation of EE compared to FEMAD. In contrast, due to large discrete state-action space and huge training work, DQN performs the slowest convergence as well as the lowest EE performance.

Fig. \ref{Fig.5} compares EE of the system assisted by MF-RIS under different configurations: MF-RIS with full energy harvesting, half energy harvesting with $\alpha_m = 0.5$, no energy harvesting with $\alpha_m = 1$, MF-RIS without signal amplification with $\beta_m = 1$, and a system without RIS. The observed trend indicates that as the number of LEO satellites increases, system EE initially improves but then declines. This suggests that while adding more LEO satellites enhances EE at first, the increasing energy consumption eventually surpasses the system's energy harvesting and management capabilities, leading to a reduction in EE. When partial energy harvesting is enabled with $\alpha_m = 0.5$, the system maintains stable performance, as MF-RIS continues to support its core functions. However, when energy harvesting is completely disabled with $\alpha_m = 1$, the system struggles to handle the operational overhead of MF-RIS, resulting in a significant drop in EE. This underscores the essential role of energy harvesting in sustaining high system performance. Additionally, when signal amplification is disabled with $\beta_m = 1$, the system experiences reduced signal gain, which lowers the transmission rate and negatively impacts energy efficiency. These results highlight the critical importance of both energy harvesting and signal amplification in optimizing system efficiency, as the absence of either functionality leads to substantial performance degradation.

Fig. \ref{Fig.6} reveals EE performance versus different numbers of MF-RIS elements $M$. Initially, EE performance escalates as the number of MF-RIS elements increases since more element can conduct signal amplification and EH functionality for enhancing high rate and lower power consumption, respectively. However, EE declines when the number of elements becomes too large, as the rising energy consumption of MF-RIS outweighs the system's energy harvesting capability. Moreover, we compare the scenarios where only a portion of the elements are randomly activated, i.e., $10\%$, $30\%$, and $50\%$ of MF-RIS elements. We can observe that the case with all elements switched on performs the highest EE performance due to its maximum flexibility. We observe that the highest EE occurs at $M=36$ in the full-on case and at $M=64$ in the half-on case. Despite having the same number of activated MF-RIS elements, the randomly selected elements in the half-on case with $M=64$ are likely to experience worse channel conditions, resulting in a lower EE. The lowest EE is observed when only $10\%$ of the MF-RIS elements are used, due to insufficient signal amplification and EH capabilities.

Fig. \ref{Fig.7} illustrates the EE performance of FEMAD with different numbers of LEOs clustered in an FL group and with different LEO transmit antennas $N$. We consider $L=10$ LEOs. Note that information exchange is conducted only within the FL group, i.e., inter-group information exchange is not applicable here. The general trend indicates that as $N$ increases, EE improves due to higher flexibility of finding better beamforming weights. Intriguingly, the optimal EE performance is accomplished when $2$ LEO agents are grouped together, with a total of $5$ FL groups in the network. This is because fewer LEO satellites in each group can more effectively adapt to similar local environmental conditions. In contrast, when more LEO satellites with greater distances are included in an FL group, they share more diverse regional information, leading to highly divergent models that reduce overall effectiveness. Additionally, the absence of information exchange, regarded as distributed learning, may result in competitive, non-cooperative policies that focus on individual optimization, ultimately yielding the lowest EE performance.

\section{Conclusions}
	This paper presents a novel architecture, LEO-MF-RIS, where MF-RIS is deployed across multiple LEO satellites to address the challenges of insufficient energy storage and long-distance signal attenuation in LEO communications. The long-term EE problem is formulated by optimizing MF-RIS configurations, including amplification, phase-shifts, energy harvesting ratios, and LEO transmit beamforming. We have designed an FEMAD scheme, where DDPG generates suitable policies with continuous parameters, and FL enables hidden information exchange to enhance performance. Numerical results demonstrate that the proposed FEMAD achieves the highest EE compared to existing methods such as centralized reinforcement learning of DQN and DDPG as well as distributed multi-agent DDPG. Additionally, the LEO-MF-RIS architecture outperforms scenarios with fixed or no energy harvesting in MF-RIS, conventional reflection-only RIS, and deployment without RIS or MF-RIS.

\linespread{0.8}
\bibliographystyle{IEEEtran}
\bibliography{IEEEabrv}

\begin{thebibliography}{10}
\providecommand{\url}[1]{#1}
\csname url@samestyle\endcsname
\providecommand{\newblock}{\relax}
\providecommand{\bibinfo}[2]{#2}
\providecommand{\BIBentrySTDinterwordspacing}{\spaceskip=0pt\relax}
\providecommand{\BIBentryALTinterwordstretchfactor}{4}
\providecommand{\BIBentryALTinterwordspacing}{\spaceskip=\fontdimen2\font plus
\BIBentryALTinterwordstretchfactor\fontdimen3\font minus
  \fontdimen4\font\relax}
\providecommand{\BIBforeignlanguage}[2]{{%
\expandafter\ifx\csname l@#1\endcsname\relax
\typeout{** WARNING: IEEEtran.bst: No hyphenation pattern has been}%
\typeout{** loaded for the language `#1'. Using the pattern for}%
\typeout{** the default language instead.}%
\else
\language=\csname l@#1\endcsname
\fi
#2}}
\providecommand{\BIBdecl}{\relax}
\BIBdecl

\bibitem{1}
L.-H. Shen, K.-T. Feng, and L.~Hanzo, ``Five facets of {6G}: Research
  challenges and opportunities,'' \emph{ACM Computing Surveys}, vol.~55,
  no.~11, pp. 1--39, 2023.

\bibitem{2}
Q.~Wu and R.~Zhang, ``Towards smart and reconfigurable environment: Intelligent
  reflecting surface aided wireless network,'' \emph{IEEE Communications
  Magazine}, vol.~58, no.~1, pp. 106--112, 2020.

\bibitem{3}
L.-H. Shen, K.-T. Feng, T.-S. Lee, Y.-C. Lin, S.-C. Lin, C.-C. Chang, and S.-F.
  Chang, ``{AI}-enabled unmanned vehicle-assisted reconfigurable intelligent
  surfaces: Deployment, prototyping, experiments, and opportunities,''
  \emph{IEEE Network}, pp. 1--1, 2024.

\bibitem{my1}
L.-H. Shen, P.-C. Wu, C.-J. Ku, Y.-T. Li, K.-T. Feng, Y.~Liu, and L.~Hanzo,
  ``{D-STAR}: Dual simultaneously transmitting and reflecting reconfigurable
  intelligent surfaces for joint uplink/downlink transmission,'' \emph{IEEE
  Transactions on Communications}, vol.~72, no.~6, pp. 3305--3322, 2024.

\bibitem{6}
W.~Wang, W.~Ni, Y.~C.~E. Hui~Tian, and R.~Zhang, ``Multi-functional
  reconfigurable intelligent surface: System modeling and performance
  optimization,'' \emph{IEEE Transactions on Wireless Communications}, vol.~23,
  no.~4, pp. 3025--3041, 2024.

\bibitem{7}
W.~Wang, L.~Gao, R.~Ding, L.~Y. J.~Lei, C.~A. Chan, and X.~Gao, ``Resource
  efficiency optimization for robust beamforming in multi-beam satellite
  communications,'' \emph{IEEE Transactions on Vehicular Technology}, vol.~70,
  no.~7, pp. 6958--6968, 2021.

\bibitem{8}
T.~S. Abdu, S.~Kisseleff, E.~Lagunas, and S.~Chatzinotas, ``Flexible resource
  optimization for {GEO} multibeam satellite communication system,'' \emph{IEEE
  Transactions on Wireless Communications}, vol.~20, no.~12, pp. 7888--7902,
  2021.

\bibitem{9}
O.~Kodheli, ``Satellite communications in the new space era: A survey and
  future challenges,,'' \emph{IEEE Communications Surveys \& Tutorials},
  vol.~23, no.~1, pp. 70--109, 2021.

\bibitem{10}
W.~U. Khan, E.~Lagunas, A.~Mahmood, S.~Chatzinotas, and B.~Ottersten,
  ``{RIS}-assisted energy-efficient {LEO} satellite communications with
  {NOMA},'' \emph{IEEE Transactions on Green Communications and Networking},
  vol.~8, no.~2, pp. 780--790, 2024.

\bibitem{11}
K.~Tekbiyik, G.~K. Kurt, A.~R. Ektı, and H.~Yanikomeroglu, ``Reconfigurable
  intelligent surfaces empowered {THz} communication in {LEO} satellite
  networks,'' \emph{IEEE Access}, vol.~10, pp. 121\,957--121\,969, 2022.

\bibitem{my2}
P.-C. Wu, L.-H. Shen, K.-T. Feng, and C.-Y. Chan, ``Federated reinforcement
  learning for multi-dual-{STAR-RIS} assisted {DFRC}-enabled multi-{BS} in
  {ISAC} systems,'' in \emph{Proc. IEEE International Conference on
  Communications (ICC)}, 2024, pp. 2986--2991.

\bibitem{13}
X.~Mu, Y.~Liu, L.~Guo, J.~Lin, and R.~Schober, ``Joint deployment and multiple
  access design for intelligent reflecting surface assisted networks,''
  \emph{IEEE Transactions on Wireless Communications}, vol.~20, no.~10, pp.
  6648--6664, 2021.

\bibitem{14}
D.~Tse and P.~Viswanath, ``Fundamentals of wireless communication. cambridge,''
  \emph{U.K. Cambridge Univ. Press}, 2005.

\bibitem{15}
P.~Zeng, Q.~Wu, and D.~Qiao, ``Energy minimization for {IRS}-aided wpcns with
  non-linear energy harvesting model,'' \emph{IEEE Communications Letters},
  vol.~10, no.~11, pp. 2592--2596, 2021.

\bibitem{thzris}
Z.~Wang, X.~Mu, J.~Xu, and Y.~Liu, ``Simultaneously transmitting and reflecting
  surface ({STARS}) for terahertz communications,'' \emph{IEEE Journal of
  Selected Topics in Signal Processing}, vol.~17, no.~4, pp. 861--877, 2023.

\bibitem{16}
Y.~Song, X.~Li, H.~Ji, and H.~Zhang, ``Energy-aware task offloading and
  resource allocation in the intelligent {LEO} satellite network,'' \emph{Proc.
  IEEE Annual International Symposium on Personal, Indoor and Mobile Radio
  Communications (PIMRC)}, pp. 481--486, 2022.

\bibitem{17}
P.-C. Hsu, L.-H. Shen, C.-H. Liu, and K.-T. Feng, ``Federated deep
  reinforcement learning for thz-beam search with limited {CSI},'' in
  \emph{Proc. IEEE Vehicular Technology Conference (VTC-Fall)}, 2022, pp. 1--6.

\bibitem{my3}
C.-Y. Chen, L.-H. Shen, K.-T. Feng, L.-L. Yang, and J.-M. Wu, ``Edge selection
  and clustering for federated learning in optical inter-{LEO} satellite
  constellation,'' in \emph{Proc. IEEE Annual International Symposium on
  Personal, Indoor and Mobile Radio Communications (PIMRC)}, 2023, pp. 1--6.

\bibitem{18}
N.~Levanon, ``Quick position determination using 1 or 2 leo satellites,''
  \emph{IEEE Transactions on Aerospace and Electronic Systems}, vol.~34, no.~3,
  pp. 736--754, 1998.

\end{thebibliography}
\end{document}